\documentclass[runningheads]{llncs}
\usepackage[T1]{fontenc}
\usepackage{graphicx}%
\usepackage{multirow}%
\usepackage{amsmath,amssymb,amsfonts}%
\usepackage{mathrsfs}%
\usepackage[title]{appendix}%
\usepackage{textcomp}%
\usepackage{manyfoot}%
\usepackage{booktabs}%
\usepackage{enumitem}
\usepackage{algorithm}%
\usepackage{algorithmicx}%
\usepackage{algpseudocode}%
\usepackage{listings}%
\usepackage[table,xcdraw]{xcolor}%
\usepackage{booktabs}%
\usepackage{adjustbox}%
\usepackage{tabularx}%
\usepackage{diagbox}%
\usepackage{caption}%
\usepackage{subcaption, soul}%
\usepackage{hyperref}
\usepackage{booktabs}

\usepackage{acronym}
\newacro{gan}[GAN]{generative adversarial network}
\newacro{vae}[VAE]{variational autoencoder}
\newacro{cnn}[CNN]{convolutional neural network}
\newacro{dire}[DIRE]{diffusion reconstruction error}
\newacro{mlp}[MLP]{multi-layer perceptron}
\newacro{clip}[CLIP]{contrastive language image pre-training}
\newacro{vlm}[VLM]{vision-language model}
\newacro{vit}[ViT]{vision transformer}
\newacro{nlp}[NLP]{natural language processing}
\newacro{cv}[CV]{computer vision}
\newacro{blip}[BLIP]{bootstrapping language image pre-training}
\newacro{vqa}[VQA]{visual question answering}
\newacro{lora}[LoRA]{low-rank adaptation}
\newacro{peft}[PEFT]{parameter-efficient fine-tuning}
\newacro{gpt}[GPT]{generative pre-trained transformer}
\newacro{q-former}[Q-Former]{querying transformer}
\newacro{sedid}[SeDID]{stepwise error for diffusion-generated image detection}
\newacro{sd}[SD]{stable diffusion}
\newacro{lsun}[LSUN]{large-scale scene understanding}
\newacro{fc}[FC]{fully-connected}
\newacro{deit}[DeiT]{data-efficient image transformers}
\newacro{ldm}[LDM]{latent diffusion model}
\newacro{adm}[ADM]{ablated diffusion model}
\newacro{ddpm}[DDPM]{denoising diffusion probabilistic models}
\newacro{iddpm}[IDDPM]{improved denoising diffusion probabilistic models}
\newacro{pndm}[PNDM]{pseudo numerical methods for diffusion models on manifolds}
\newacro{srm}[SRM]{spatial rich model}
\newacro{lasted}[LASTED]{language-guided synthesis detection}
\newacro{rf}[RF]{random forest}
\newacro{dm}[DM]{diffusion model}
\newacro{ddim}[DDIM]{denoising diffusion implicit models}
\newacro{multilid}[multiLID]{multi local intrinsic dimensionality}
\newacro{ifdl}[IFDL]{image forgery detection and localization} 
\newacro{svm}[SVM]{support vector machine} 
\newacro{ai}[AI]{artificial intelligence} 

\newacro{amsff}[AMSFF]{attention-based multi-scale feature fusion} 
\newacro{psm}[PSM]{patch selection module}
\newacro{llm}[LLM]{large language model}
\newacro{clip}[CLIP]{Contrastive Language-Image Pretraining}
\newacro{c2p}[C2P]{category common prompt}
\newacro{sid}[SID]{synthetic image detection}
\newacro{fosid}[FOSID]{Fact-checked Online Synthetic Image Dataset}
\newacro{rasid}[RASID]{Retrieval-Assisted Synthetic Image Detection}
\newacro{gff}[GFF]{Guided and Fused Frozen CLIP-ViT}
\newacro{fuseformer}[FuseFormer]{Multi-Stage Fusion Module}
\newacro{dfgm}[DFGM]{Deepfake-Specific Feature Guidance Module}
\newacro{ffaa}[FFAA]{Face Forgery Analysis Assistant}
\newacro{mllm}[MLLM]{ Multimodal Large Language Model}
\newacro{mids}[MIDS]{Multi-answer Intelligent Decision System}
\newacro{ssm}[SSM]{state space model}
\newacro{einfft}[EinFFT]{Einstein FFT}
\newacro{fft}[FFT]{Fast Fourier Transform}
\newacro{vim}[ViM]{Vision Mamba}
\newacro{ms2d}[MS2D]{Multi-Scale 2D}
\newacro{convffn}[ConvFFN]{onvolutional Feed-Forward
Network}
\newacro{cnn}[CNN]{convolutional neural network}
\newacro{dvae}[dVAE]{discrete variational autoencoder}
\newacro{coglm}[CogLM]{cross-modal general language model}
\newacro{qformer}[Q-Former]{querying transformer}
\newacro{sdgs}[SDGS]{Synthetic Data Generation System}
\newacro{sota}[SOTA]{state-of-the-art}
\newacro{mse}[MSE]{mean squared error}
\newacro{ssim}[SSIM]{structural similarity index measure}
\newacro{psnr}[PSNR]{peak signal-to-noise ratio}
\newacro{vssd}[VSSD]{visual state space duality}
\definecolor{darkgreen}{RGB}{10,191,10}

\definecolor{olivegreen}{RGB}{212,232,231}

\definecolor{lightblue}{RGB}{138,170,229}
\colorlet{lightblueAlpha}{lightblue!30}

\definecolor{lightred}{RGB}{255, 194, 194}
\begin{document}
\title{DeeCLIP: A Robust and Generalizable Transformer-Based Framework for Detecting AI-Generated Images}
\titlerunning{DeeCLIP}
%

\author{Mamadou Keita\inst{1}\orcidID{0009-0009-7618-9253} \and
Wassim Hamidouche\inst{2}\orcidID{0000-0002-0143-1756} \and
Hessen Bougueffa Eutamene\inst{1}\orcidID{0009-0009-0556-9996} \and Abdelmalik Taleb-Ahmed\inst{1}\orcidID{0000-0001-7218-3799} \and Abdenour Hadid\inst{3}\orcidID{0000-0001-9092-735X}}
\authorrunning{M. Keita {\it et al.}}
%
\institute{Laboratory of IEMN, Univ. Polytechnique
Hauts-de-France, Valenciennes, France \and
Khalifa University, Abu Dhabi, UAE \and
Sorbonne Center for Artificial Intelligence, Sorbonne University, Abu Dhabi, UAE}
\maketitle              

\begin{abstract}
This paper introduces {\bf DeeCLIP}, a novel framework for \underline{\bf de}t\underline{\bf e}cting AI-generated images using \underline{\bf CLIP}-ViT and fusion learning. Despite significant advancements in generative models capable of creating highly photorealistic images, existing detection methods often struggle to generalize across different models and are highly sensitive to minor perturbations. To address these challenges, DeeCLIP incorporates DeeFuser, a fusion module that combines high-level and low-level features, improving robustness against degradations such as compression and blurring. Additionally, we apply triplet loss to refine the embedding space, enhancing the model’s ability to distinguish between real and synthetic content. To further enable lightweight adaptation while preserving pre-trained knowledge, we adopt parameter-efficient fine-tuning using \ac{lora} within the CLIP-ViT backbone. This approach supports effective zero-shot learning without sacrificing generalization. Trained exclusively on 4-class ProGAN data, DeeCLIP achieves an average accuracy of 89.00\% on 19 test subsets composed of \ac{gan} and diffusion models. Despite having fewer trainable parameters, DeeCLIP outperforms existing methods, demonstrating superior robustness against various generative models and real-world distortions. 
The code is publicly available at \href{https://github.com/Mamadou-Keita/DeeCLIP} {GitHub} for research purposes.
\end{abstract}

\begin{figure}[t]
      \centering
        \includegraphics[width=0.8\linewidth]{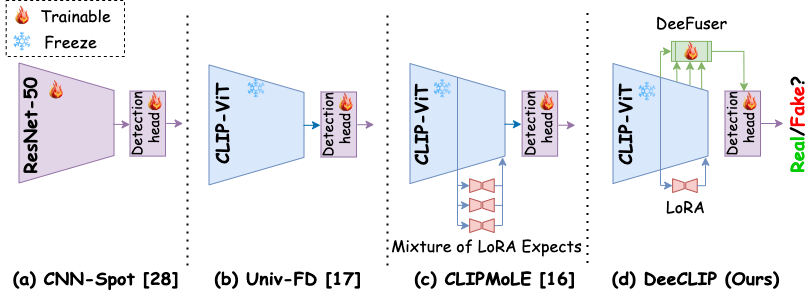}
        \caption{Fine-tuning approaches for AI-generated image detection. Compared to (a) full fine-tuning and (b) linear probing, our approach allows more effective and efficient adaption of CLIP-ViT for AI-generated image detection tasks. }
      \label{fig:teaser}  \vspace{-3mm}
\end{figure}


\vspace{-1cm}
\section{Introduction}
\label{sec:intro}

The emergence of advanced image generation models, driven by deep learning, has fundamentally transformed computer vision. In particular, \ac{gan}~\cite{goodfellow2014generative} and diffusion-based architectures~\cite{sohl2015deep} have achieved remarkable success in generating photorealistic images that closely resemble real-world visuals. While this progress has enabled new applications in entertainment, art, and content creation, it raises significant concerns regarding trust, security, and authenticity. \acs{ai}-generated images can be weaponized for malicious purposes, including disinformation, identity forgery, and privacy violations. The rapid evolution of image-generation techniques further exacerbates these risks, making robust detection methods an urgent necessity. Despite ongoing research, \acs{ai}-generated images continue to evade detection, particularly when images are synthesized by unseen models during training or when subjected to post-processing manipulations. Existing detectors often fail to generalize effectively, leading to performance degradation on new generative models. Recent efforts have explored pre-trained \ac{vlm} for this task, fine-tuning them with a detection head (see Figue~\ref{fig:teaser}). A notable example is the work by Cozzolino \textit{et al.}~\cite{cozzolino2023raising}, which employs CLIP features with a linear \ac{svm} classifier. However, these methods struggle to maintain robustness against unseen generative models and real-world distortions such as compression, and blurring. To address these limitations, we introduce DeeCLIP, a robust and generalizable transformer-based model for detecting \acs{ai}-generated images. DeeCLIP builds upon the contrastively pre-trained CLIP-ViT model, leveraging its open-world visual knowledge and extensive exposure to diverse image-text pairs. Current CLIP-based detection approaches~\cite{ojha2023towards,cozzolino2023raising} primarily rely on deep features extracted from the final or penultimate layer of the visual encoder. However, this approach overlooks crucial shallow-level details, such as pixel-level inconsistencies and texture artifacts, which are often present in AI-generated images. While multi-layer feature fusion has proven effective in other vision tasks~\cite{lin2017feature}, its application in AI-generated image detection remains underexplored~\cite{cao2024mmfuser}. To bridge this gap, DeeCLIP introduces the DeeFuser module, a modified version of MMFuser~\cite{cao2024mmfuser}, which dynamically integrates deep and shallow features. This multi-scale fusion enhances fine-grained representations, improving robustness against compression artifacts, adversarial perturbations, and novel generative models. Additionally, DeeCLIP is trained end-to-end with triplet loss, refining the learned embedding space for better separation between authentic and AI-generated images. To maintain the generalization strength of the pre-trained CLIP model, we adopt parameter-efficient fine-tuning via \acs{lora}. This allows DeeCLIP to adapt effectively with minimal adjustment, without overwriting CLIP’s pre-trained knowledge. Our extensive experiments demonstrate that DeeCLIP achieves \ac{sota} performance in cross-generator generalization and robustness to image perturbations. Notably, trained solely on a subset (4-class) of ProGAN data, DeeCLIP attains an average accuracy of 89.00\% on 19 test subsets composed of \ac{gan} and diffusion models. Furthermore, DeeCLIP exhibits strong robustness to image degradation, achieving a higher overall average accuracy of 71.91\% across all degradation types compared to 61.55\% for the \ac{sota} model C2P-CLIP, a gain of 10.36\%. This robustness is particularly evident under challenging conditions such as Gaussian blur and JPEG compression, validating its effectiveness in real-world scenarios. Further, we evaluate 4-class ProGAN-trained DeeCLIP on an entirely different dataset to assess its generalization capability. It still achieves a solid average accuracy of \textbf{78.99\%}, highlighting its adaptability to previously unseen data distributions. Our key contributions are summarized as follows:

\begin{itemize}[label=\textbullet]
\item \textbf{Robust Multi-Scale Feature Fusion:} DeeCLIP incorporates the DeeFuser module, which dynamically integrates shallow and deep features, ensuring the detection of subtle pixel-level artifacts and high-level inconsistencies in AI-generated images.
\item \textbf{Improved Generalization via Triplet Loss:} Triplet loss refines the embedding space, enhancing the separation of real and AI-generated images. This structured separation improves robustness against unknown generative models and real-world perturbations.

\item \textbf{Prompt-Free Generalization:} While some previous methods, such as DIRE~\cite{wang2023dire}, also explore prompt-free generalization, DeeCLIP achieves this naturally without relying on text-based prompts while integrating feature fusion for stronger robustness.
\item \textbf{State-of-the-Art Robustness and Generalization:} DeeCLIP achieves 89.00\% average accuracy on diverse GAN and diffusion subsets and outperforms prior methods under distortions, with a 10.36\% gain over C2P-CLIP. It also generalizes well, achieving 78.99\% accuracy on a different dataset with real and synthetic data.
\end{itemize}

\begin{figure*}[t]
    \centering
    \includegraphics[width=.7\linewidth]{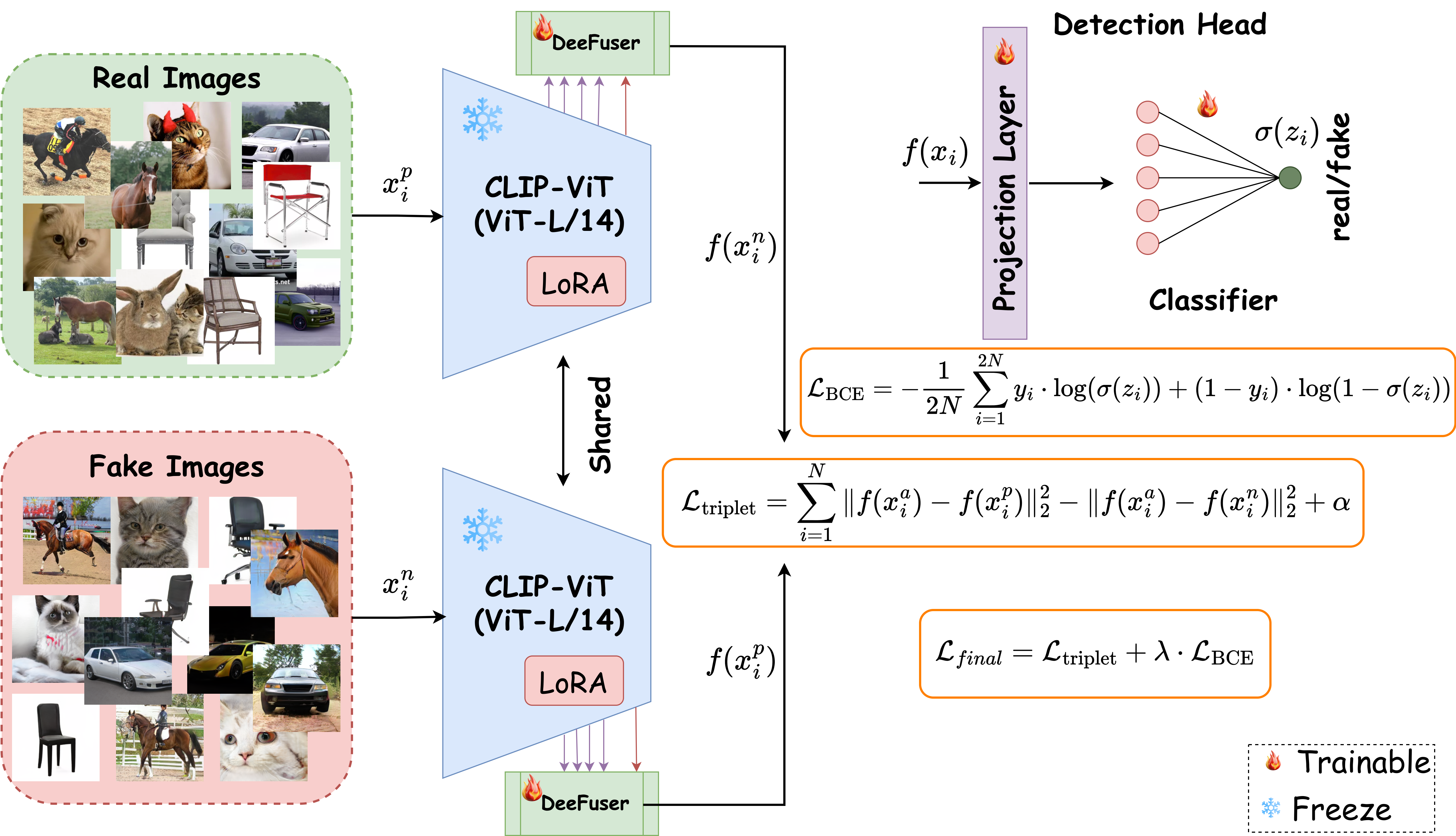}
    \caption{DeeCLIP's architecture. Given an image, deep and shallow features are extracted using CLIP-ViT (ViT-L/14) fine-tuned with LoRA. Then, the DeeFuser module fuses these features, capturing both semantic meaning and fine-grained textures. The fused representation is processed through a projection layer, where the output is fed into the classifier to distinguish real from fake images.}
    \label{fig:DeeCLIP}  \vspace{-3mm}
\end{figure*}

\vspace{-5mm}
\section{Proposed Approach: DeeCLIP}
\label{sec:method}
\vspace{-2mm}
The overall framework of our approach is shown in Figure~\ref{fig:DeeCLIP}. It comprises three core blocks: the CLIP-ViT model as the backbone, a deep and shallow feature fusion module called DeeFuser, and a linear classifier for AI-generated image detection. We refer to this framework as DeeCLIP.\\

\noindent {\bf CLIP-ViT Encoder:}
To leverage the substantial visual-world knowledge and efficient feature extraction capabilities of CLIP-ViT, we adapt its image encoder as the backbone feature extractor for our approach. While one might consider fine-tuning all or part of the parameters, this method may result in performance degradation for two primary reasons: first, fine-tuning large pre-trained CLIP parameters on a limited dataset risks overfitting; second, fine-tuning could distort CLIP’s pre-existing visual-world knowledge, leading to reduced performance under significant distribution shifts. 
Inspired by recent advances in parameter-efficient transfer learning, we propose training task-specific adapter modules while keeping the pre-trained CLIP parameters frozen to preserve its valuable knowledge. To achieve this, we adopt \ac{lora}~\cite{hu2021lora}, a widely used method in \ac{nlp} and computer vision tasks. \Ac{lora} enhances transfer learning without imposing substantial computational overhead by modifying the model's transformer blocks. Instead of directly altering the weight matrix \(W\), LoRA decomposes the weight update \(\Delta W\) into two low-rank matrices \(A\) and \(B\), thus reducing computational cost and preserving the original weight matrix \(W\). The adapted weight is then expressed as \(W' = W + BA\), ensuring minimal disruption to the pre-trained parameters.

Given an input image \(x \in \mathcal{R}^{3 \times h \times w}\), where \(h\) and \(w\) represent the height and width of the image, the patch embedding layer divides the image into patches. These patches are subsequently transformed into a sequence of embeddings \(E_{img} \in \mathcal{R}^{(N+1) \times D}\), where \(N = \frac{h \cdot w}{P^2}\) is the number of patches, \(P\) is the patch size, and \(D\) is the dimensionality of the embeddings. The sequence includes a CLS token (indicated by the 1st term) that represents the entire image. This sequence of embedded patches is then processed through multiple Vision Transformer (ViT) blocks, and the output token of each patch is used to form the image encoding vector \(f_i \in \mathcal{R}^{N \times D}\), where \(f_i\) represents the output at the \(i\)-th stage of the CLIP-ViT network.\\

\begin{figure*}[t]
    \centering
    \begin{subfigure}[b]{0.49\linewidth}
        \includegraphics[width=\linewidth]{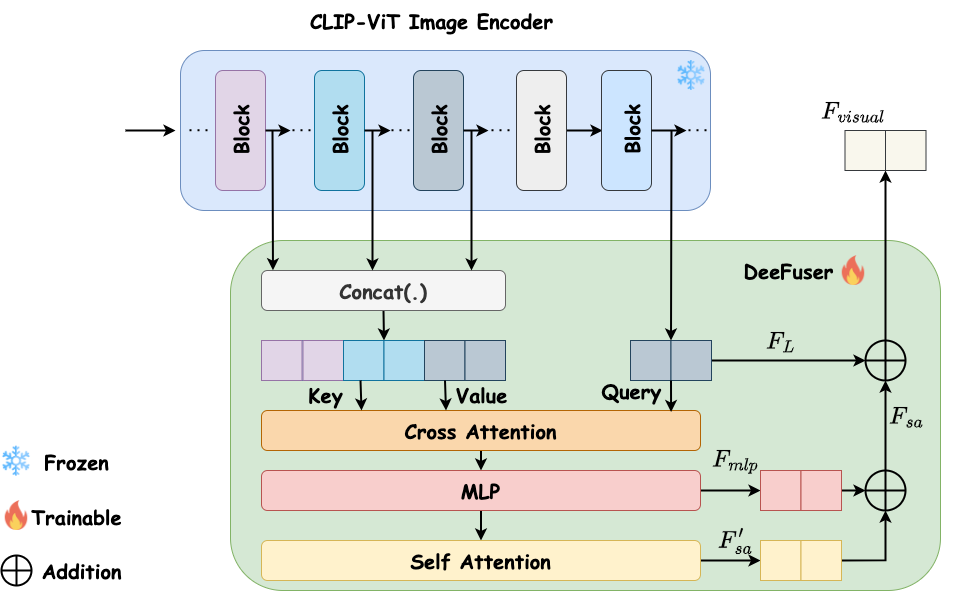}
        \caption{DeeFuser Module}\label{subfig:deefuser}
    \end{subfigure}
   \vspace{.001in}
    \begin{subfigure}[b]{0.49\linewidth}\includegraphics[width=\linewidth]{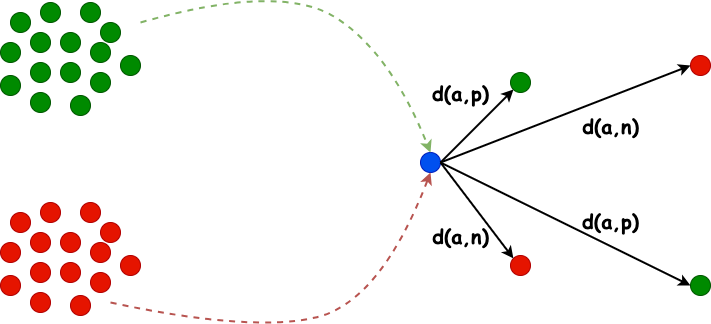}
        \caption{Triplet Loss} \label{subfig:triplet}
    \end{subfigure}
    \caption{(a) DeeFuser module to fuse  both deep and shallow features from CLIP-ViT encoder. (b) Illustration of triplet loss: the anchor sample (blue) is randomly selected from either the positive (green) or negative (red) samples. If the anchor is from the positive class, it is optimized to be closer to other positive samples, further away from negative samples, and vice versa.}
    \label{fig:DeeFuser}  \vspace{-3mm}
\end{figure*}

\noindent {\bf DeeFuser Module:}
After extracting features using CLIP-ViT, we intend to leverage low-level and high-level features to explore its potential fully. Build on the assumption that the shallow and deep features of CLIP-ViT can complement each other. We leverage a multi-layer feature fusion module, MMFuser~\cite{cao2024mmfuser}, that we slightly modify. DeeFuser serves as a bridge between the CLIP-ViT vision encoder and the linear classifier. The DeeFuser architecture is illustrated in Figure~\ref{subfig:deefuser}. Concretely, we extract $L$ feature maps from the CLIP-ViT image encoder, denoted as \( F = [F_1, F_2, \dots, F_L] \). Then, we use the deep feature \( F_L \) as a query to dynamically capture missing details from the shallow-level feature maps \( X = \text{Concat}(F_1, F_2, \dots, F_{L-1}) \) through a cross-attention operation. This results in a refined visual feature representation \( F_{ca} \in \mathbb{R}^{N \times D} \) with enriched fine-grained details, which can be expressed as:

\[
F_{ca} = \text{Attention}(\text{norm}(F_L), \text{norm}(X)),
\] where \( \text{Attention}(\cdot) \) denotes the attention mechanism, \( \text{norm}(\cdot) \) represents layer normalization, and \( \text{Concat}(\cdot) \) is the concatenation operation. Following this, we apply a multi-layer perceptron (MLP) to \( F_{ca} \) to enhance its expressive power by adding non-linearity and further refining feature relationships. The output of the MLP, denoted as \( F_{mlp} \), is formulated as:

\[
F_{mlp} = \text{MLP}(\text{norm}(F_{ca})),
\] where the MLP consists of two fully connected layers with Gelu activation functions, which project \( F_{ca} \) into a transformed space. This layer captures more complex interactions within the feature map and enhances discriminative properties. To further emphasize relevant features, we introduce a self-attention mechanism into \( F_{mlp} \) as follows:

\[
F'_{sa} = \text{Attention}(\text{norm}(F_{mlp}), \text{norm}(F_{mlp})),
\]
\[
F_{sa} = F_{mlp} + \alpha_2 F'_{sa},
\] where \( \alpha_2 \in \mathbb{R}^{D} \) is a learnable parameter adjusting the influence of \( F_{mlp} \) relative to \( F'_{sa} \). Finally, we combine the refined feature map \( F_{sa} \) with the original deep feature \( F_L \) using another learnable vector \( \alpha_1 \in \mathbb{R}^{D} \), resulting in the final enhanced visual feature \( F_{\text{visual}} \):
\[
F_{\text{visual}} = F_L + \alpha_1 F_{sa}.
\] This enhanced visual feature \( F_{\text{visual}} \) integrates finer, multi-scale information and serves as a superior input for the linear classifier, facilitating more accurate detection.\\

\noindent {\bf Loss functions:}
For the loss calculation, we utilize the triplet loss~\cite{weinberger2009distance}, as depicted in Figure~\ref{subfig:triplet}. The triplet loss aids the model in learning more discriminative features by encouraging it to place similar images closer together in the embedding space, while pushing dissimilar images further apart. Specifically, we randomly select a triplet of images: an anchor image, which can be either a real or a fake sample. Depending on whether the anchor is real or fake, we then choose a positive image (from the same category as the anchor) and a negative image (from the opposite category). The objective is to minimize the distance between the anchor and the positive image while maximizing the distance between the anchor and the negative image. This process helps the model learn robust features that can effectively distinguish real samples from fake ones.

The triplet loss is computed as follows:
\begin{equation*}
    \mathcal{L}_{\text{triplet}} = \sum_{i=1}^{N}  \| f(x_i^a) - f(x_i^p) \|_2^2 - \| f(x_i^a) - f(x_i^n) \|_2^2 + \alpha
\end{equation*}

where \( N \) is the number of triplets in the batch, \( x_i^a \), \( x_i^p \), and \( x_i^n \) represent the anchor, positive, and negative samples in the \( i \)-th triplet, respectively. \( f(x) \) denotes the embedding output from the DeeFuser module for a given image \( x \), and \( \| \cdot \|_2 \) represents the Euclidean distance between two embeddings. The term \( \alpha \) is the margin, a constant that specifies the minimum distance between the anchor-positive and anchor-negative pairs.

For the detection loss, we adopt binary cross-entropy with logits loss, which is computed as follows:
\begin{equation*}
    \mathcal{L}_{\text{BCE}} = - \frac{1}{2N} \sum_{i=1}^{2N}  y_i \cdot \log(\sigma(z_i)) + (1 - y_i) \cdot \log(1 - \sigma(z_i))
\end{equation*}

where \( y_i \) represents the ground truth label for the \( i \)-th sample, with \( y_i = 0 \) for real samples and \( y_i = 1 \) for fake samples. \( z_i \) denotes the raw, unnormalized logits output from the model for the \( i \)-th sample, and \( \sigma(z) \) refers to the sigmoid function applied to the logits \( z \), converting them into probabilities.

Finally, the total model loss is defined as the sum of the triplet loss and the detection loss, weighted by a scaling factor \( \lambda \):
\begin{equation*}
    \mathcal{L}_{\text{final}} = \mathcal{L}_{\text{triplet}} + \lambda \cdot \mathcal{L}_{\text{BCE}}
\end{equation*}

where \( \lambda \) is a hyperparameter used to balance the contributions of the two loss components.

\begin{table*}[t]
\caption{Accuracy (ACC) scores of state-of-art detectors and DeeCLIP across 19 test datasets. Best performance is denoted with \textbf{bold}.}
\label{tab:TrainedOnUniversalFakeDetect}

\begin{adjustbox}{width=\linewidth}
\begin{tabular}{@{}l|c|ccccccccccccccccccc|c@{}}
\toprule
\multirow{2}{*}{Method} & \multirow{2}{*}{Ref} & \multicolumn{6}{c}{GAN} & \multirow{2}{*}{\begin{tabular}[c]{@{}c@{}}Deep\\ Fakes\end{tabular}} & \multicolumn{2}{c}{Low level} & \multicolumn{2}{c}{Perceptual loss} & \multirow{2}{*}{Guided} & \multicolumn{3}{c}{LDM} & \multicolumn{3}{c}{Glide} & \multirow{2}{*}{Dalle} & \multirow{2}{*}{mAcc} \\ \cmidrule(lr){3-8} \cmidrule(lr){10-13} \cmidrule(lr){15-20}
\multicolumn{1}{c|}{} &  & \begin{tabular}[c]{@{}c@{}}Pro-\\ GAN\end{tabular} & \begin{tabular}[c]{@{}c@{}}Cycle-\\ GAN\end{tabular} & \begin{tabular}[c]{@{}c@{}}Big-\\ GAN\end{tabular} & \begin{tabular}[c]{@{}c@{}}Style-\\ GAN\end{tabular} & \begin{tabular}[c]{@{}c@{}}Gau-\\ GAN\end{tabular} & \begin{tabular}[c]{@{}c@{}}Star-\\ GAN\end{tabular} &  & SITD & SAN & CRN & IMLE &  & \begin{tabular}[c]{@{}c@{}}200\\ steps\end{tabular} & \begin{tabular}[c]{@{}c@{}}200\\ w/cfg\end{tabular} & \begin{tabular}[c]{@{}c@{}}100\\ steps\end{tabular} & \begin{tabular}[c]{@{}c@{}}100\\ 27\end{tabular} & \begin{tabular}[c]{@{}c@{}}50\\ 27\end{tabular} & \begin{tabular}[c]{@{}c@{}}100\\ 10\end{tabular} &  &  \\ \cmidrule(r){1-22} 

Freq-spec & WIFS 2019 & 49.90 & \bf99.90 & 50.50 & 49.90 & 50.30 & 99.70 & 50.10 & 50.00 & 48.00 & 50.60 & 50.10 & 50.90 & 50.40 & 50.40 & 50.30 & 51.70 & 51.40 & 50.40 & 50.00 & 55.45 \\

Co-occurence & Elect. Imag. & 97.70 & 97.70 & 53.75 & 92.50 & 51.10 & 54.70 & 57.10 & 63.06 & 55.85 & 65.65 & 65.80 & 60.50 & 70.70 & 70.55 & 71.00 & 70.25 & 69.60 &  69.90 & 67.55 & 66.86 \\

CNN-Spot & CVPR 2020 & 99.99 & 85.20 & 70.20 & 85.70 & 78.95 & 91.70 & 53.47 & 66.67 & 48.69 & 86.31 & 86.26 & 60.07 & 54.03 & 54.96 & 54.14 & 60.78 & 63.80 & 65.66 & 55.58 & 69.58 \\


Patchfor & ECCV 2020 & 75.03 & 68.97 & 68.47 & 79.16 & 64.23 & 63.94 & 75.54 & 75.14 & 75.28 & 72.33 & 55.30 & 67.41 & 76.50 & 76.10 & 75.77 & 74.81 & 73.28 & 68.52 & 67.91 & 71.24 \\

F3Net & ECCV 2020 & 99.38 & 76.38 & 65.33 & 92.56 & 58.10 & \bf100.00 & 63.48 & 54.17 & 47.26 & 51.47  & 51.47  & 69.20 &  68.15 & 75.35 & 68.80  &  81.65 & 83.25 & 83.05  & 66.30 & 71.33  \\



Bi-LORA & ICASSP 2023 & 98.71 & 96.74 & 81.18 & 78.30 & 96.30 & 86.32 & 57.78 & 68.89 & 52.28 & 73.00 & 82.60 & 65.10 & 85.15 & 59.20 & 85.00 & 83.50 & 85.65 & 84.90 & 72.70 & 78.59 \\

LGrad & CVPR 2023 & 99.84 & 85.39 & 82.88 & 94.83 & 72.45 & 99.62 & 58.00 & 62.50 & 50.00 & 50.74 & 50.78 & 77.50 & 94.20 & 95.85 & 94.80 & 87.40 & 90.70 & 89.55 & 88.35 & 80.28 \\

UniFD & CVPR 2023 & \bf100.00 & 98.50 & 94.50 & 82.00 & \bf99.50 & 97.00 & 66.60 & 63.00 & 57.50 & 59.50 & 72.00 & 70.03 & 94.19 & 73.76 & 94.36 & 79.07 & 79.85 & 78.14 & 86.78 & 81.38 \\

AntiFakePrompt & CVPR 2023 & 99.26 & 96.82 & 87.88 & 80.00 & 98.13 & 83.57 & 60.20 & 70.56 & 53.70 & 79.21 & 79.01 & 73.75 & 89.55 & 64.10 & 89.80 & 93.55 & 93.90 & 92.95 & 80.10 & 82.42 \\

FreqNet & AAAI 2024 & 97.90 & 95.84 & 90.45 & 97.55 & 90.24 & 93.41 & 97.40 & 88.92 & 59.04 & 71.92 & 67.35 & \bf86.70 & 84.55 & \bf99.58 & 65.56 & 85.69 & \bf97.40 & 88.15 & 59.06 & 85.09 \\


NPR & CVPR 2024 & 99.84 & 95.00 & 87.55 & 96.23 & 86.57 & 99.75 & 76.89 & 66.94 & \bf98.63 & 50.00 & 50.00 & 84.55 & 97.65 & 98.00 & 98.20 & \bf96.25 & 97.15 & \bf97.35 & 87.15 & 87.56 \\

FatFormer & CVPR 2024 & 99.89 & 99.32 & \bf99.50 & 97.15 & 99.41 & 99.75 & 93.23 & 81.11 & 68.04 & 69.45 & 69.45 & 76.00 & 98.60 & 94.90 & 98.65 & 94.35 & 94.65 & 94.20 & 98.75 & 90.86 \\

RINE & ECCV 2024 & {\bf 100.00} & 99.30 & 99.60 & 88.90 & 99.80 & 99.50 & 80.60 & 90.60 & 68.30 & 89.20 & 90.60 & 76.10 & 98.30 & 88.20 & 98.60 & 88.90 & 92.60 & 90.70 & 95.00 & 91.31\\

C2P-CLIP$^\star$& Arxiv 2024& 99.71 & 90.69 & 95.28 & \bf99.38 & 95.26 & 96.60 & 89.86 & \bf98.33 & 64.61 & 90.69 & 90.69 & 77.80 & 99.05 & 98.05 & 98.95 & 94.65 & 94.20 & 94.40 & \bf98.80 & 93.00 \\

C2P-CLIP$^\ddagger$& Arxiv 2024& 99.98 & 97.31 & 99.12 & 96.44 & 99.17 & 99.60 & \bf93.77 & 95.56 & 64.38 & \bf93.29 & 93.29 & 69.10 & \bf99.25 & 97.25 & \bf99.30 & 95.25 & 95.25 & 96.10 & 98.55 & \bf93.79 \\

\rowcolor{lightblueAlpha}  DeeCLIP (ours) & - & {\bf 100.00} & 97.69 & 98.32 & 97.81 & 94.87 & 99.97 & 62.31 & 84.72 & 57.53 & 90.18 & 90.18 & 77.20 &  98.90 & 98.50 & 98.80 & 80.50 & 82.85 & 82.40 & 98.35 & 89.00 \\

\bottomrule
\end{tabular}
\end{adjustbox}
\begin{flushleft}
($\star$) Trump,Biden. ($\ddagger$)  Deepfake,Camera.
\end{flushleft}
\vspace{-6mm}
\end{table*}
\vspace{-3mm}
\section{Experimental Results and Analysis}
\label{sec:experiments}
\vspace{-3mm}

\noindent \textbf{Dataset.} To ensure a fair comparison, we utilize the widely recognized UniversalFakeDetect dataset~\cite{ojha2023towards}, which has been extensively used in prior benchmarks. This allows for a direct evaluation of DeeCLIP against \ac{sota} methods, ensuring consistency and robustness in performance assessment. Following the experimental setup introduced by Wang {\it et al.}~\cite{wang2020cnn}, the dataset employs ProGAN as the training set, comprising 20 subsets of generated images. For training, we adopt only 4-class setting (horse, chair, cat, car) as outlined by Tan {\it et al.}~\cite{tan2024c2p}. The test set consists of 19 subsets generated by a diverse range of models, such as ProGAN, StyleGAN, BigGAN, CycleGAN, StarGAN, and GauGAN~\cite{karras2019style}, Deepfake~\cite{rossler2019faceforensics}, CRN, IMLE, SAN, and SITD~\cite{li2019diverse,dai2019second}, as well as Guided Diffusion, LDM, GLIDE, and DALLE~\cite{dhariwal2021diffusion,rombach2022high}.

To further evaluate DeeCLIP's generalization capability, we compare it against the best-performing models trained on the ProGAN 4-class setup using a different testing dataset. This dataset includes 12 subsets: 2 real data subsets (MS COCO and Flickr) and 10 synthetic subsets (ControlNet, DALL·E 3, DiffusionDB, IF, LaMA, LTE, SD2Inpaint, SDXL, SGXL, and SD3).   

\noindent {\textbf{\\Evaluation Metrics.}} Following the convention of previous detection methods~\cite{keita2024bi,tan2024c2p}, we report the accuracy (ACC). We also calculate the mean accuracy across all data subsets to provide a more comprehensive evaluation of overall model performance.\\

\noindent {\textbf{Baselines.}} 
In our study, we fine-tuned AntiFakePrompt~\cite{chang2023antifakeprompt} and Bi-LORA~\cite{keita2024bi}. Moreover, we have chosen the latest and most competitive methods in the field, including, frequency-based methods: Co-occurence, Freq-spec, FreqNet, F3Net~\cite{tan2024frequency}, CNN-based methods: CNN-Spot, FatchFor, NPR~\cite{wang2020cnn,tan2024rethinking}, transformer-based methods: UniFD, C2P-CLIP, Fatformer~\cite{liu2024forgery}, and LGrad~\cite{tan2023learning}. For all those models, we report the results presented in the original C2P-CLIP~\cite{tan2024c2p} paper. Finally, for RINE, we report results from its paper~\cite{koutlis2024leveraging}.\\

\noindent\textbf{Implementation Details.} We fine-tune all layers of the CLIP-ViT (ViT-L/14) image encoder using LoRA, setting the rank to 16, the alpha to 32, and applying a dropout rate of 0.05. The training utilizes the AdamW optimizer with a learning rate of  $5\cdot10^{-5}$. A batch size of 8 is used, and the model is trained for 5 epochs. The fusion process selects features from 12 specific layers of the ViT model: 1, 3, 5, 8, 10, 13, 15, 17, 19, 21, 22, and 23. Among these, the first 11 layers (1 to 22) are low-level features, while the last layer (23) represents high-level features. The loss scaling factor ($\lambda$) is set to 2.0. For both training and testing, we follow the pre-processing pipeline specified in the original CLIP model implementation. All experiments were conducted on a system equipped with an NVIDIA RTX A4500 GPU (16 GB) and an Intel(R) i9-12950HX CPU running Ubuntu 22.04.5 LTS.\\

\noindent\textbf{Comparison with SOTA.}
Table~\ref{tab:TrainedOnUniversalFakeDetect} provides a comprehensive comparison of the cross-model detection performance of our approach and various baselines on the UniversalFakeDetect dataset, demonstrating its effectiveness in detecting images from multiple types of generative models. UniversalFakeDetect testing set includes images from 19 different generative models, covering \ac{gan}, deepfakes, low-level vision models, perceptual loss models, and diffusion models. DeeCLIP achieves a competitive mean accuracy (mAcc) of 89.00\%, showing good cross-model detection ability and making it one of the best-performing methods in this comprehensive evaluation.

On \acs{gan}-based generative models, DeeCLIP achieves outstanding performance, with a mean accuracy of 98.11\%. These results highlight the model's strong capability to distinguish between various \acs{gan} architectures, surpassing most baseline methods within this category. For diffusion-based models, DeeCLIP demonstrates excellent generalization, achieving high accuracy across different configurations: 98.35\% for Glide, 99.8\% for LDM\_100\_steps, 98.9\% for LDM\_200\_steps, and an average accuracy of 90\% across all diffusion-based generative models. While models like FatFormer, RINE, and C2P-CLIP show marginally better performance in certain cases, DeeCLIP consistently maintains a high level of accuracy. In challenging scenarios, such as deepfakes, low-level, and perceptual loss models, DeeCLIP remains competitive, outperforming baselines like UniFD and LGrad. This balanced performance across diverse image generation methods underscores DeeCLIP’s robustness and practical applicability.
\begin{table*}[t]
\caption{Generalization performance of methods trained on 4-class ProGAN. Results show accuracy (\%) on real and synthetic data subsets, each containing 3,000 image samples.}
\label{tab:generalization}
\begin{adjustbox}{width=\linewidth}
\begin{tabular}{l|c|cc|cccccccccc|c}
\toprule
Methods & \#params & MS COCO & Flickr & ControlNet & Dalle3 & DiffusionDB & IF & LaMA & LTE & SD2Inpaint & SDXL & SGXL & SD3 & mAcc \\ \midrule
FatFormer & 493M & 33.97 & 34.04 & 28.27 & 32.07 & 28.10 & 27.95 & 28.67 & 12.37 & 22.63 & 31.97 & 22.23 & 35.91 & 28.18\\ 
RINE & 434M & \bf 99.80 & \bf 99.90 & \bf 91.60 & 75.00 & \bf 73.00 & 77.40 & 30.90 & 98.20 & 71.90 & 22.20 & 98.50 & 08.30 & 70.56\\ 
 C2P-CLIP & 304M & 99.67 & 99.73 & 15.10 & \bf 75.57 & 27.87 & \bf 89.56 & \bf 65.43 & 00.20 & 27.90 & \bf 82.90 & 07.17 & \bf 70.46 & 55.13\\ 
\rowcolor{lightblueAlpha}  DeeCLIP (ours) & 306M & 97.83 & 98.50 & 86.03 & 69.33 & 71.10 & 61.37 & 63.07 & \bf 99.97 & \bf 80.57 & 62.60 & \bf98.90 & 58.61 & \bf 78.99\\ \bottomrule
\end{tabular}
\end{adjustbox}  \vspace{-3mm}
\end{table*}

\begin{figure*}[t]
    \centering
    \begin{subfigure}[b]{0.24\linewidth}
        \includegraphics[width=\linewidth]{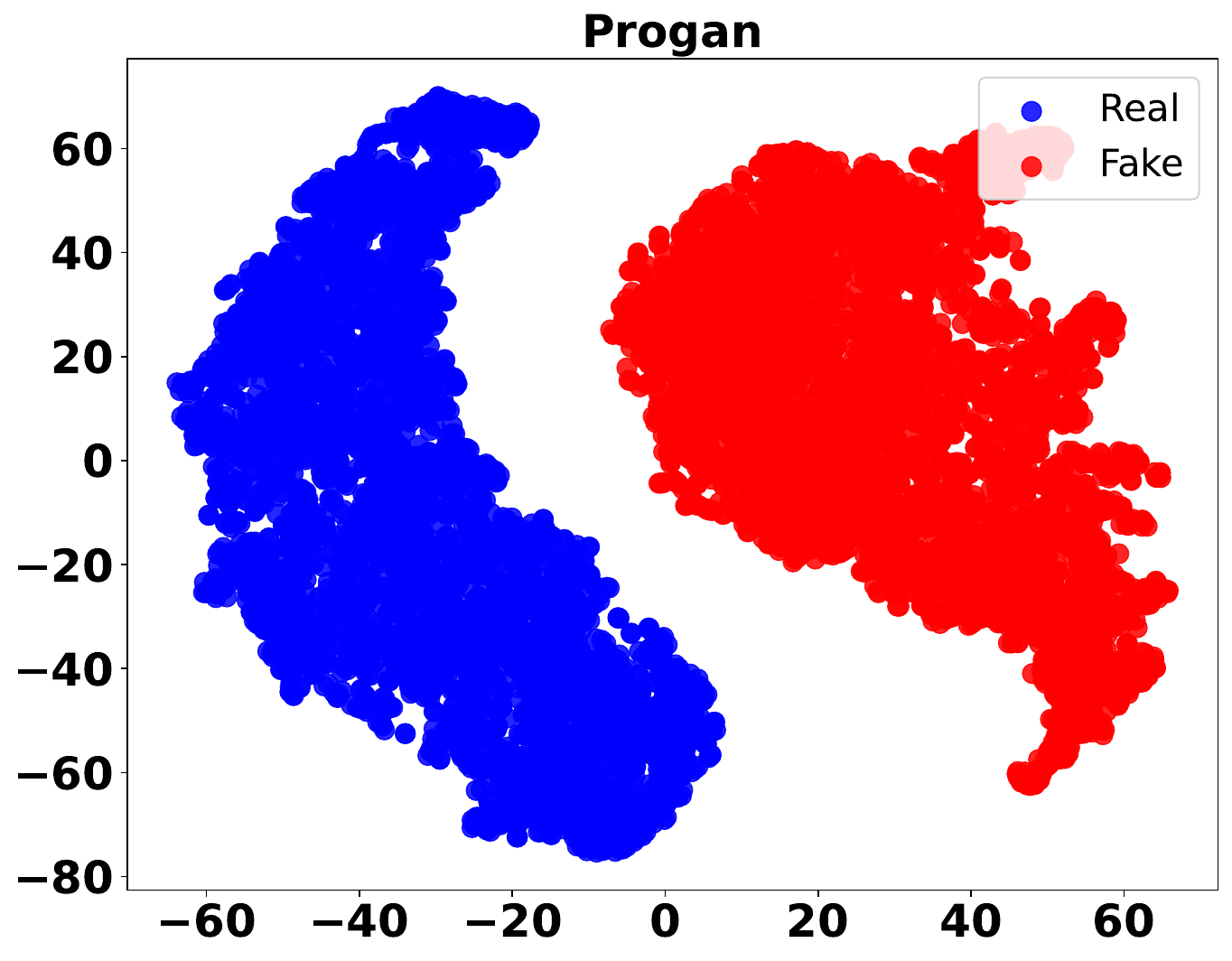}
    \end{subfigure}
   \vspace{.001in}
    \begin{subfigure}[b]{0.24\linewidth}
        \includegraphics[width=\linewidth]{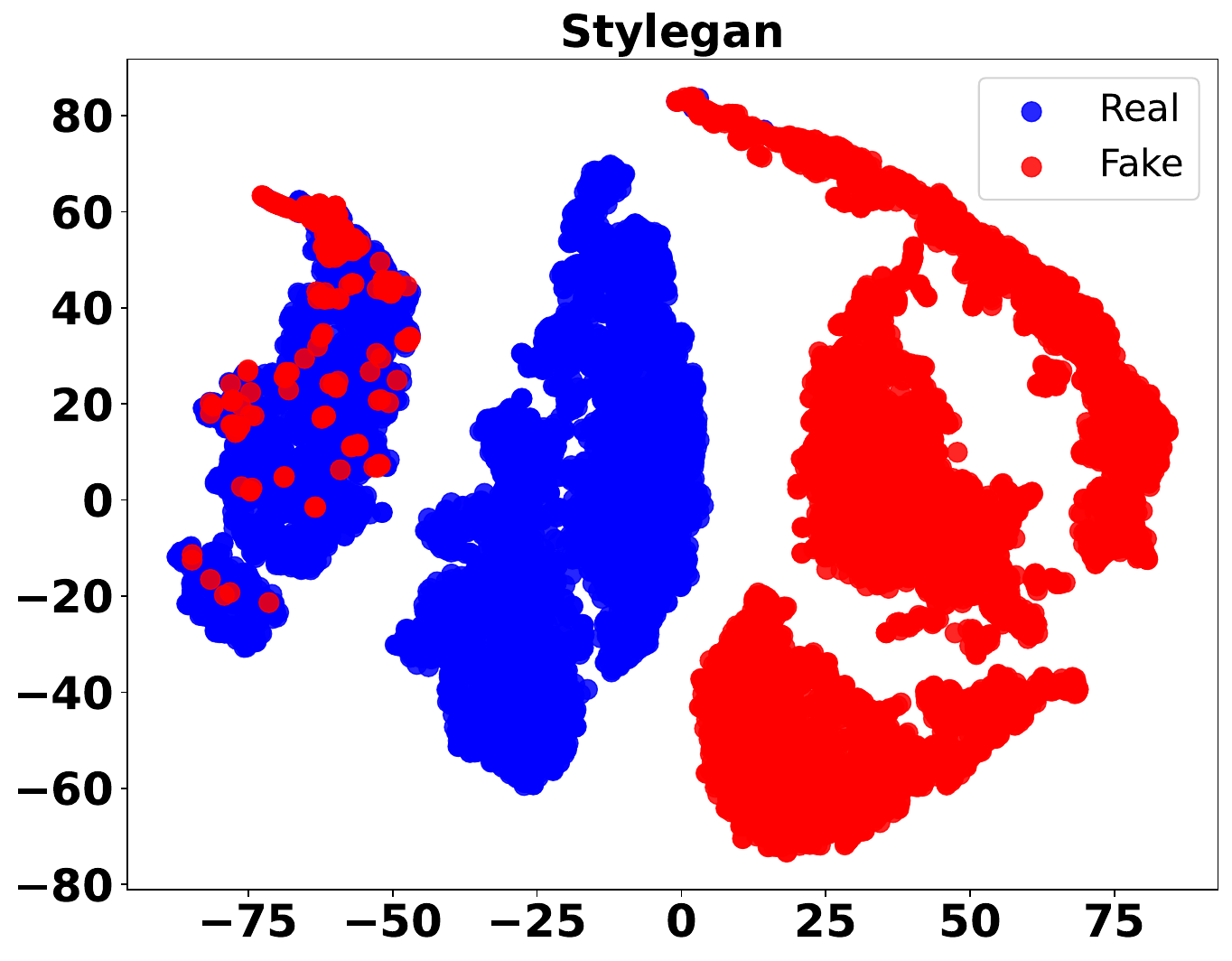}
    \end{subfigure}
   \vspace{.001in}
    \begin{subfigure}[b]{0.24\linewidth}
        \includegraphics[width=\linewidth]{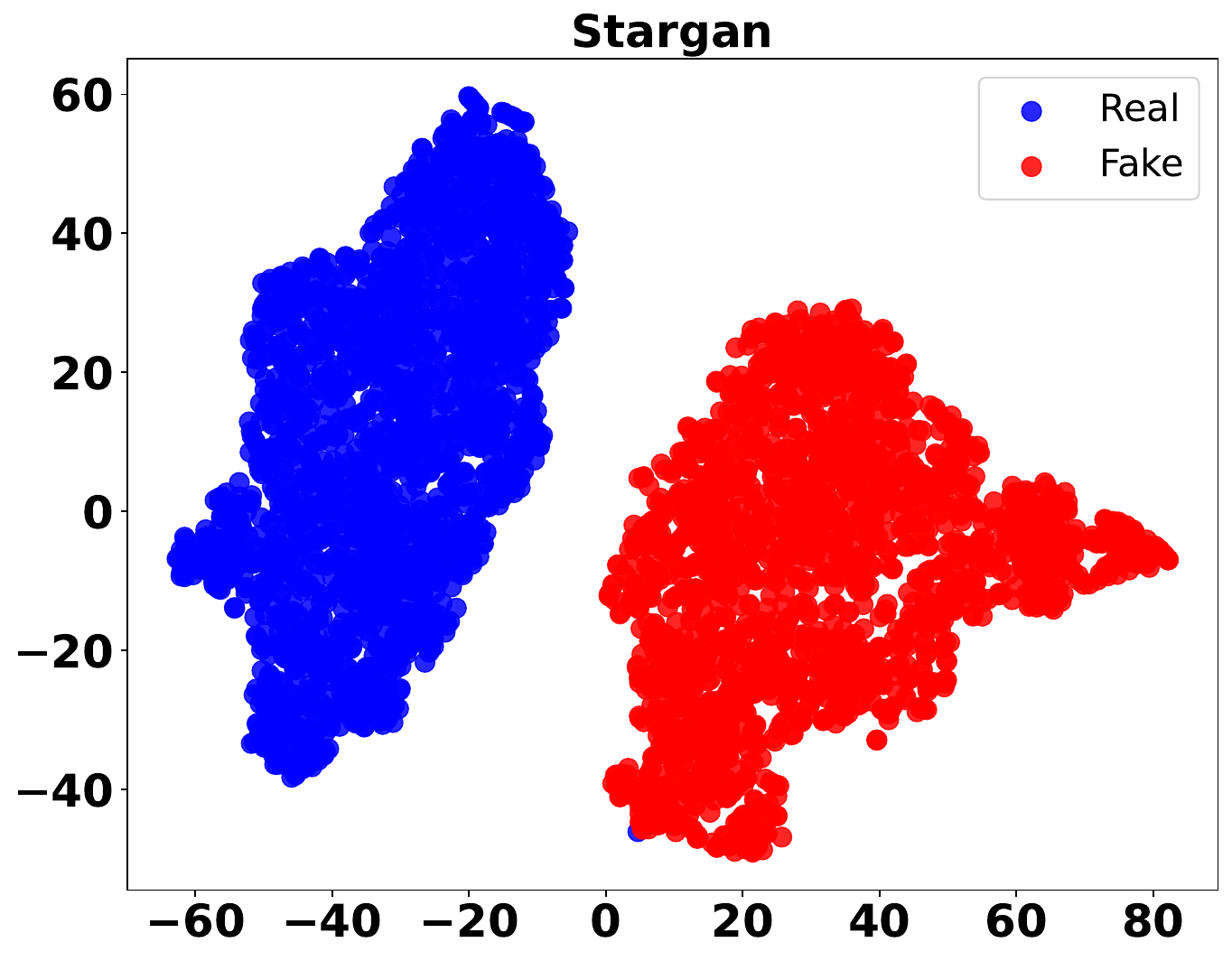}
    \end{subfigure}
   \vspace{.001in}
    \begin{subfigure}[b]{0.24\linewidth}
        \includegraphics[width=\linewidth]{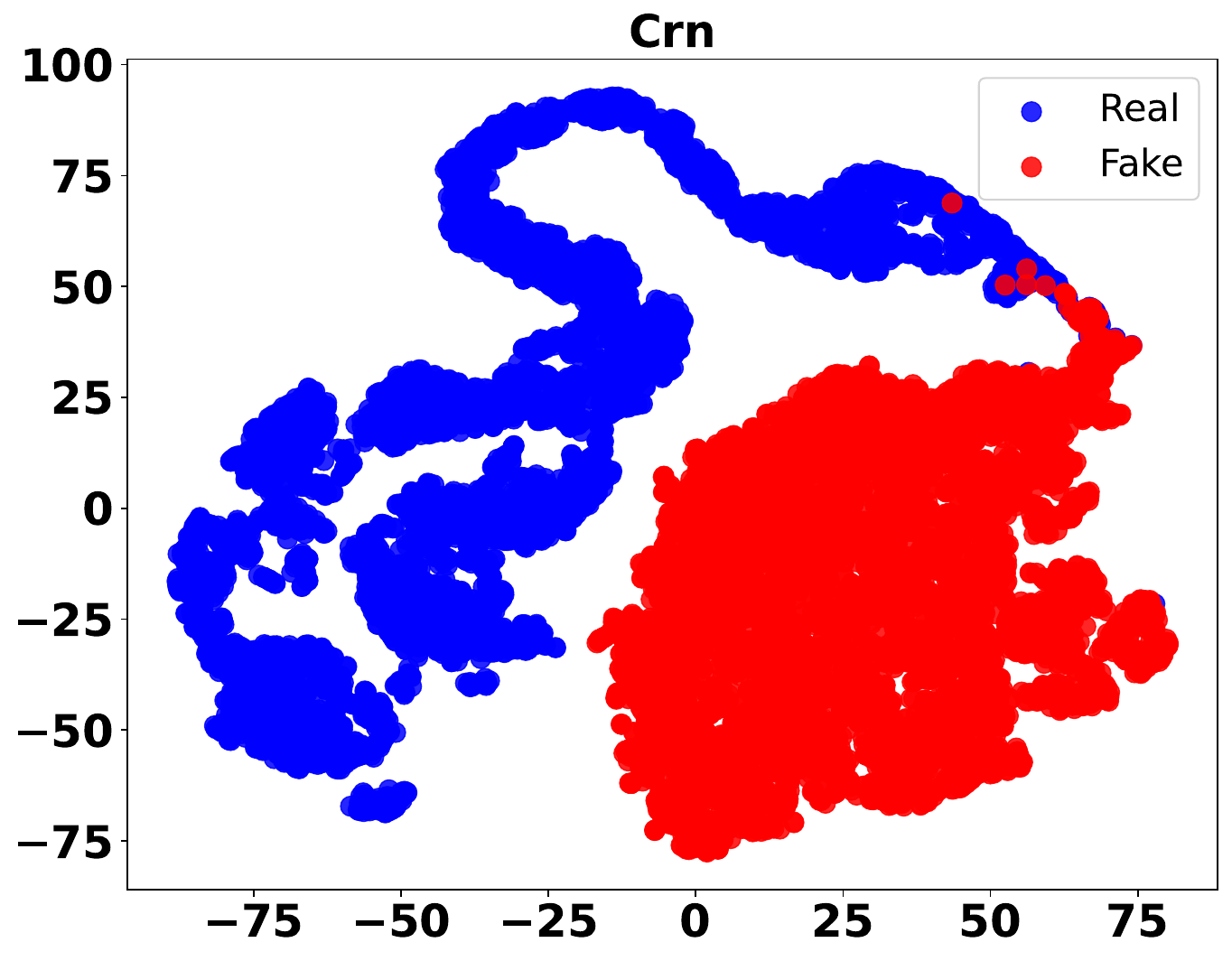}
    \end{subfigure}
    \begin{subfigure}[b]{0.24\linewidth}
\includegraphics[width=\linewidth]{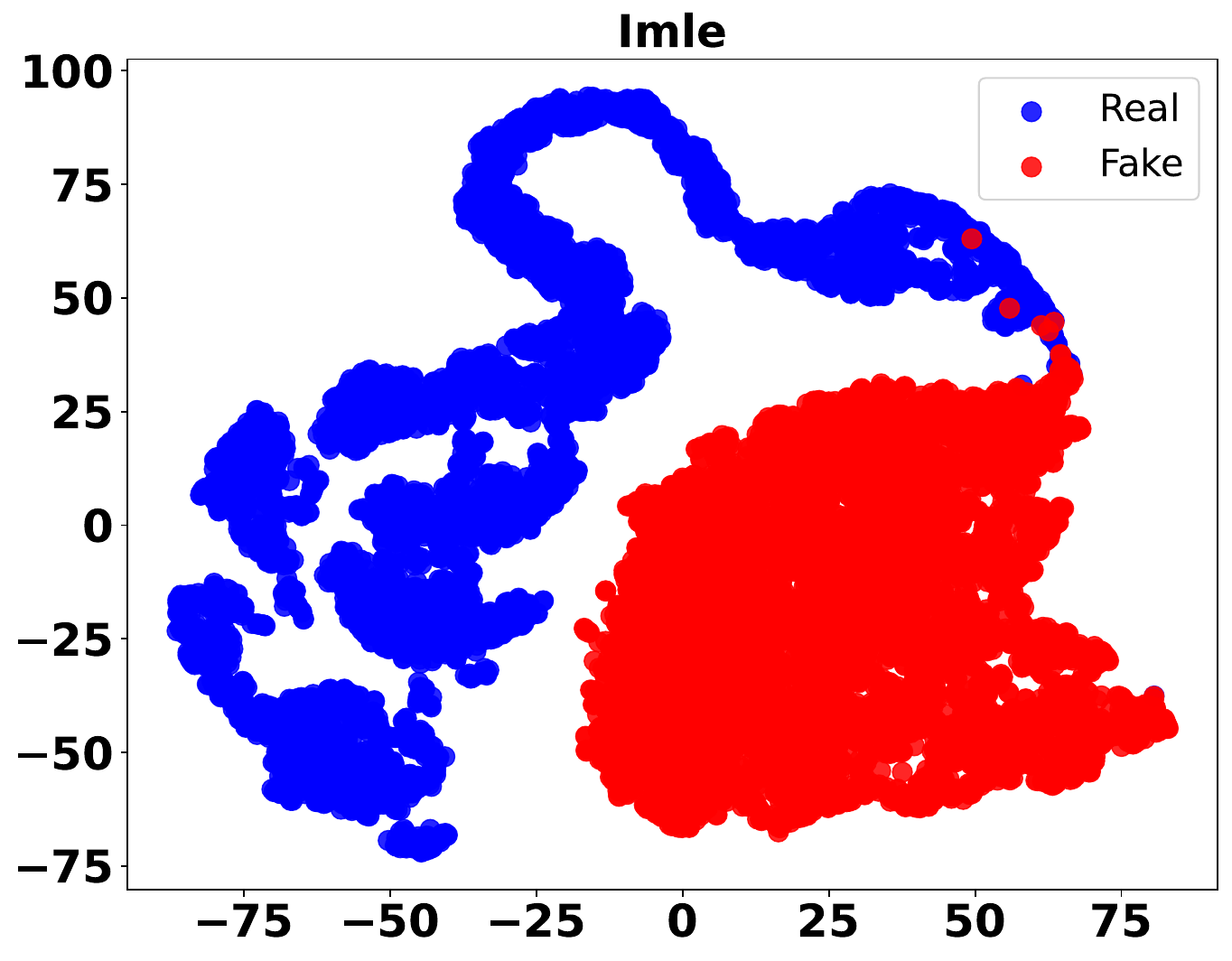}
    \end{subfigure}
   \vspace{.001in}
    \begin{subfigure}[b]{0.24\linewidth}
        \includegraphics[width=\linewidth]{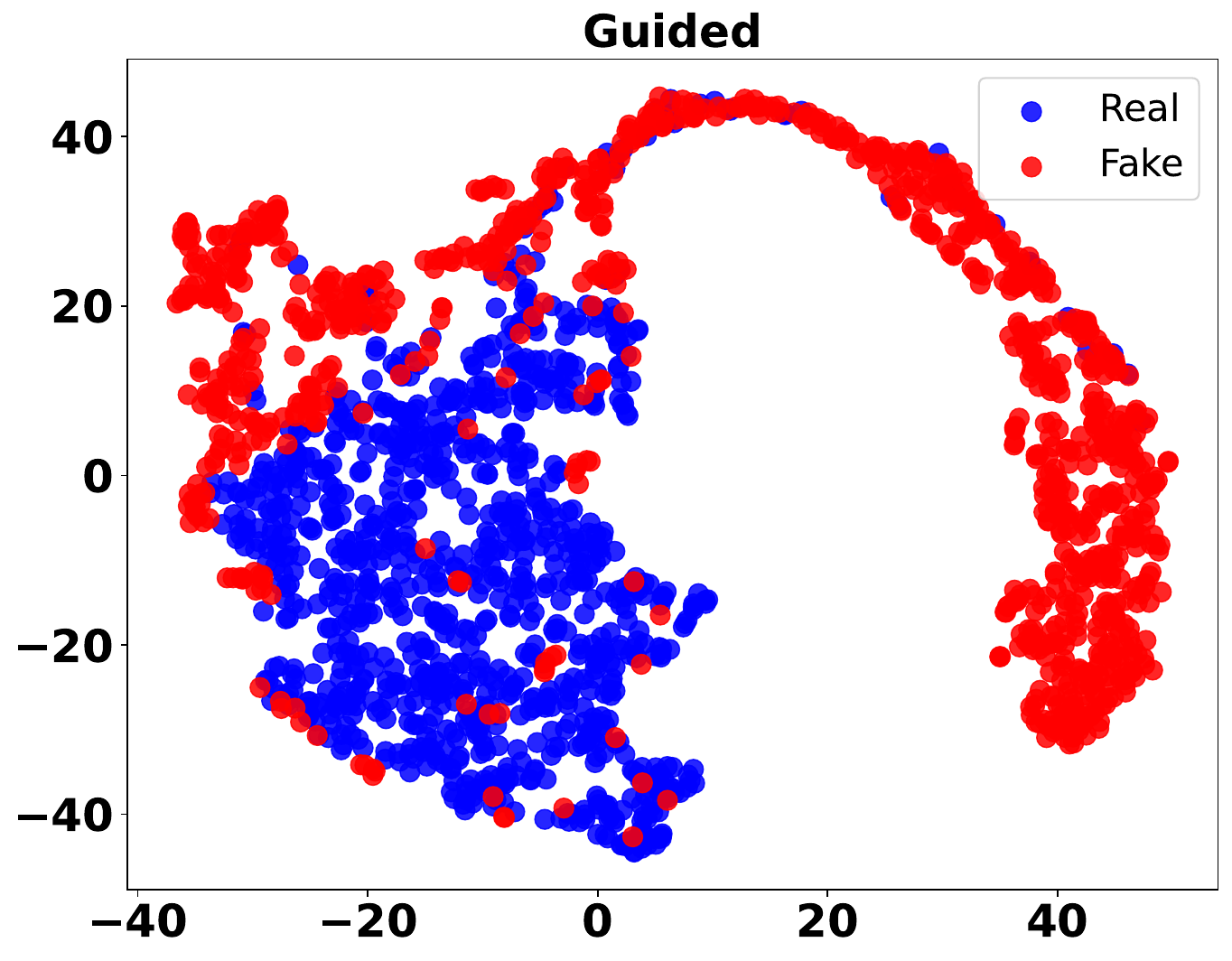}
    \end{subfigure}
    \vspace{.001in}
    \begin{subfigure}[b]{0.24\linewidth}     \includegraphics[width=\linewidth]{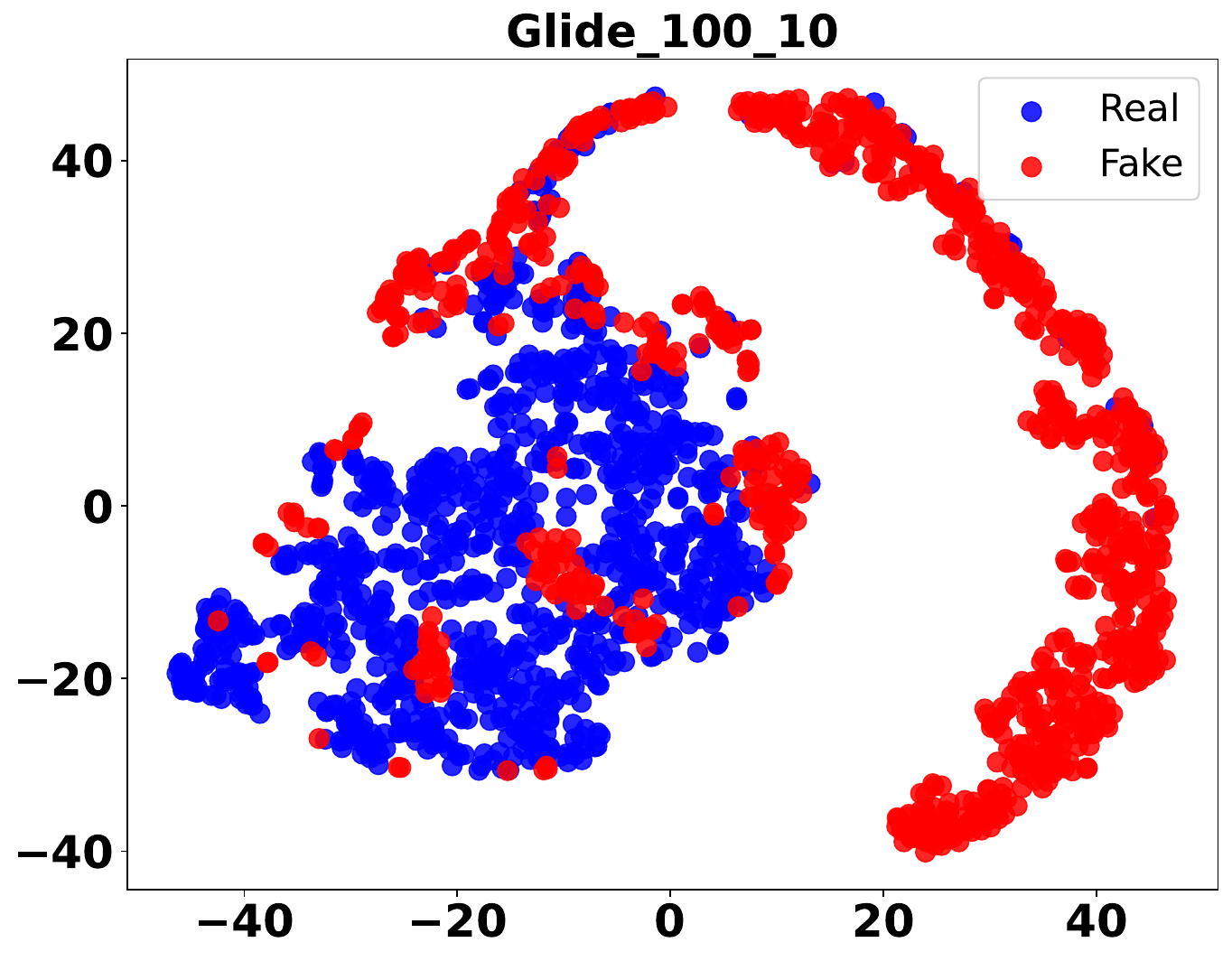}
    \end{subfigure}
   \vspace{.001in}
     \begin{subfigure}[b]{0.24\linewidth}     \includegraphics[width=\linewidth]{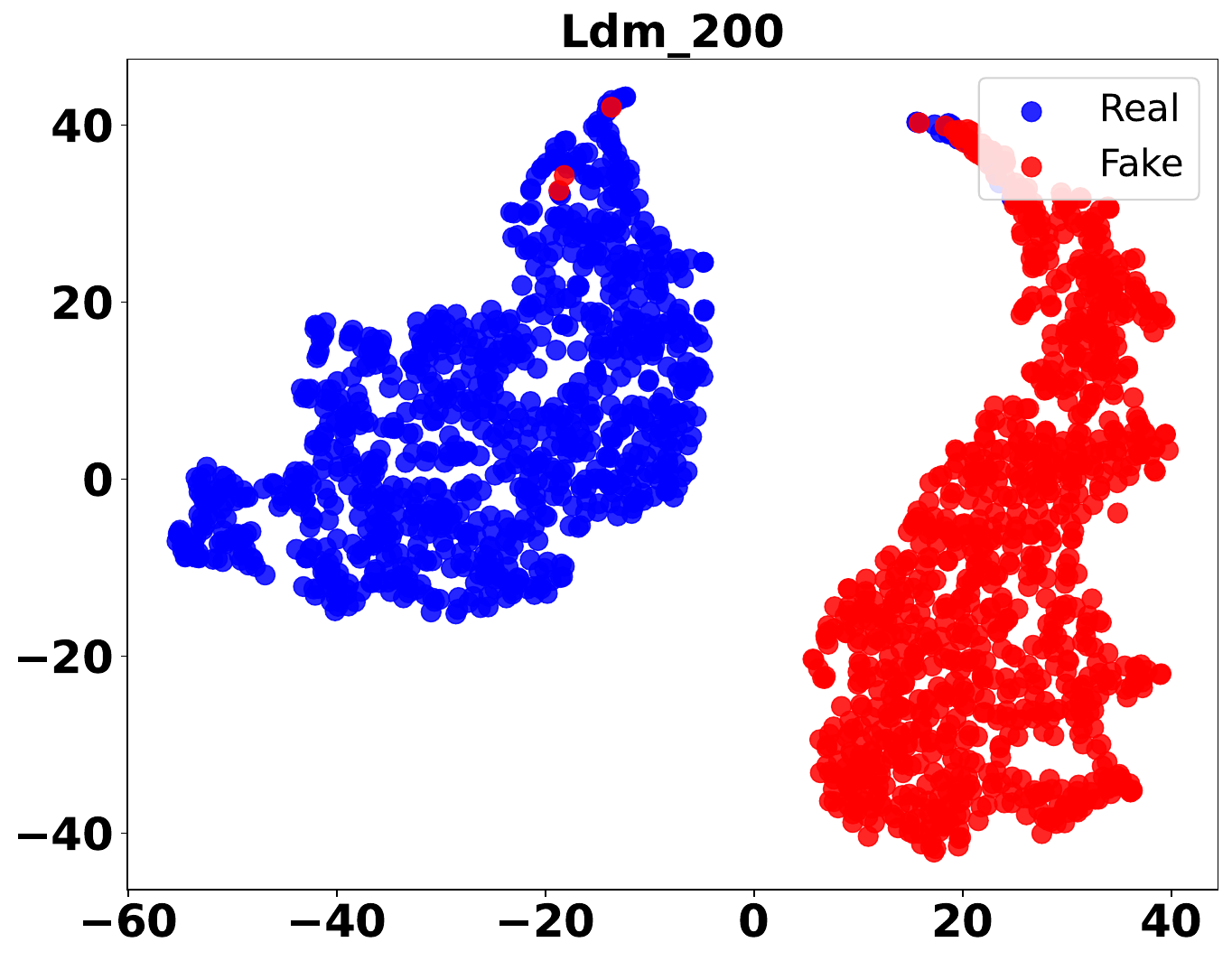}
     \end{subfigure}  
    \caption{t-SNE visualization of feature distributions for different generative models. The scatter plots illustrate the t-SNE embeddings of features extracted from real (green) and generated (red) images across various generative models, showing how well the features separate real from fake images.}
    \label{fig:Visual}   \vspace{-3mm}
\end{figure*}
Furthermore, unlike prompt-based approaches (i.e., AntiFakePrompt~\cite{chang2023antifakeprompt}), DeeCLIP achieves these results without relying on task-specific text prompts, enhancing its versatility in detecting images from a wide range of generative models. Our findings suggest that the combination of \ac{lora}-based adaptation and a deep-shallow feature fusion strategy significantly enhances CLIP-ViT’s ability to generalize to unseen generative models. These results affirm that DeeCLIP not only competes with \ac{sota} AI-generated image detectors but also offers a scalable and efficient solution for detecting images from various generative models. Figure~\ref{fig:Visual} illustrates that DeeCLIP effectively differentiates features from distinct classes, with authentic images represented in green and fake images in red. The visualizations demonstrate DeeCLIP’s capability to capture generalizable features, enabling it to adapt to unseen data distributions.\\

\noindent\textbf{DeeCLIP's Generalization Capability.}
 We evaluated the generalization capabilities of the four best-performing methods from Table~\ref{tab:TrainedOnUniversalFakeDetect}, FatFormer, RINE, C2P-CLIP, and DeeCLIP, when trained on the ProGAN 4-class setup and tested on diverse real-world generative models. The testing subsets include real data (MS COCO and Flickr) and synthetic data (ControlNet, DALL·E 3, DiffusionDB, and others). This setup assesses how well each method generalizes to unseen data distributions, particularly those differing from the training domain.

Table~\ref{tab:generalization} presents a comparative evaluation of the generalization performance of these methods on real and synthetic data. DeeCLIP consistently emerges as the most robust model, demonstrating strong generalization capabilities across both categories. It achieves near-perfect accuracy on real datasets (e.g., 98.50\% on Flickr) and excels on synthetic subsets, such as LTE (99.97\%) and SGXL (98.90\%). RINE also performs well but exhibits variability in synthetic data detection, with a notable drop in SD3 (8.30\%). C2P-CLIP shows mixed performance, excelling on some synthetic subsets (e.g., 89.56\% on IF) while struggling significantly on others (e.g., 0.20\% on LTE). FatFormer, however, consistently underperforms, highlighting its limitations in adapting to new data distributions.\\ 

\begin{table*}[t]
\caption{Accuracy Performance Under Common Image Degradations: JPEG Compression and Gaussian Blur. Best average performance is denoted with \textbf{bold}.}
\label{tab:perturb}
\begin{adjustbox}{width=\textwidth}
\begin{tabular}{@{}l|c|ccccccccccccccccccc|c@{}}
\toprule
\multirow{2}{*}{Method} & \multirow{2}{*}{Degradation} & \multicolumn{6}{c}{GANs} & \multirow{2}{*}{\begin{tabular}[c]{@{}c@{}}Deep\\ fake\end{tabular}} & \multicolumn{2}{c}{Low level} & \multicolumn{2}{c}{Perceptual loss} & \multirow{2}{*}{Guided} & \multicolumn{3}{c}{LDM} & \multicolumn{3}{c}{Glide} & \multirow{2}{*}{Dalle} & \multirow{2}{*}{mAcc} \\ \cmidrule(lr){3-7} \cmidrule(lr){10-13} \cmidrule(lr){15-20}
 & & \begin{tabular}[c]{@{}c@{}}Pro-\\ GAN\end{tabular} & \begin{tabular}[c]{@{}c@{}}Cycle-\\ GAN\end{tabular} & \begin{tabular}[c]{@{}c@{}}Big-\\ GAN\end{tabular} & \begin{tabular}[c]{@{}c@{}}Style-\\ GAN\end{tabular} & \begin{tabular}[c]{@{}c@{}}Gau-\\ GAN\end{tabular} & \begin{tabular}[c]{@{}c@{}}Star-\\ GAN\end{tabular} &  & SITD & SAN & CRN & IMLE &  & \begin{tabular}[c]{@{}c@{}}200\\ steps\end{tabular} & \begin{tabular}[c]{@{}c@{}}200\\ w/cfg\end{tabular} & \begin{tabular}[c]{@{}c@{}}100\\ steps\end{tabular} & \begin{tabular}[c]{@{}c@{}}100\\ 27\end{tabular} & \begin{tabular}[c]{@{}c@{}}50\\ 27\end{tabular} & \begin{tabular}[c]{@{}c@{}}100\\ 10\end{tabular} &  &  \\ \cmidrule(r){1-22} 

C2P-CLIP & Jpeq \(q= 80\) &  95.80 &  94.93 &  92.92 &  75.60 &  96.85 & 95.02 &  85.57 &  94.72 &  55.94 &  92.53 &  92.42 & 64.80 & 14.30 & 53.10 & 16.80 &  58.40 & 63.40 &  56.80 & 17.10 & 69.32\\

C2P-CLIP &  Jpeq \(q= 70\) & 94.49 & 94.44 & 87.10 & 65.30 & 95.18 & 92.72 & 84.90 & 93.89 & 54.79 & 86.45 & 88.09 &  70.30 & 24.10 & 67.70 & 27.90 & 61.70 &  64.10 & 56.70 & 30.50 &  70.54 \\

C2P-CLIP &  Jpeq \(q= 60\) & 94.59 & 94.40 & 81.80 & 62.30 & 95.57 & 89.62 & 82.76 & 90.56 & 53.65 & 80.00 & 80.05 & 65.90 & 23.20 &  69.70 & 26.10 & 58.00 & 60.50 & 54.50 & 42.10 & 68.70 \\ 

\rowcolor{lightblueAlpha}  DeeCLIP (ours) &  Jpeq \(q= 80\) & 90.34 & 92.01 & 81.88 & 55.82 & 90.99 & 90.67 & 50.77 & 81.94 & 51.14 & 55.81 & 60.25 & 53.40 &  70.70 & 57.55 &  71.25 & 55.35 & 54.30 & 54.60 & 79.50 & 68.33 \\ 

\rowcolor{lightblueAlpha}  DeeCLIP (ours) &  Jpeq \(q= 70\) & 85.90 & 85.39 & 73.22 & 52.54 & 87.97 & 79.31 & 50.66 & 84.44 & 50.23 & 52.08 & 55.86 & 54.05 & 69.30 & 56.05 & 70.05 & 55.95 & 54.55 & 55.00 & 75.40 & 65.68 \\

\rowcolor{lightblueAlpha}  DeeCLIP (ours) &  Jpeq \(q= 60\) & 79.79 & 76.87 & 63.75 & 50.98 & 81.22 & 65.53 & 50.31 & 76.67 & 50.00 & 50.97 & 52.13 & 53.35 & 65.95 & 54.20 & 67.20 & 54.40 & 53.35 & 53.85 & 68.45 & 61.52 \\  \midrule
C2P-CLIP &  Blur \(\sigma =1\) & 96.10 & 90.31 & 97.02 &  97.00 &  95.75 & 96.80 &93.43 & 95.56 & 57.08 & 68.84 & 68.84 & 47.90 & 01.20 & 06.60 & 01.60 & 12.50 & 13.30 & 10.60 & 01.60 & 55.37 \\ 
C2P-CLIP  &  Blur \(\sigma =2\) & 72.20 & 85.24 & 87.35 & 79.45 & 90.08 & 86.47 & 80.33 & 95.56 & 51.60 & 60.90 & 61.17 & 45.40 & 01.20 & 06.60 & 01.60 & 12.50 & 13.30 & 10.60 & 03.50 & 49.74 \\
C2P-CLIP &  Blur \(\sigma =3\) & 77.20 & 84.18 & 77.03 & 62.87 & 87.19 & 88.64 & 75.52 & 95.00 & 49.54 & 56.39 & 56.39 & 47.80 & 13.50 & 35.30 & 12.20 & 42.00 & 40.40 & 42.20 & 13.70 & 55.63 \\ 
\rowcolor{lightblueAlpha}  DeeCLIP (ours) &  Blur \(\sigma =1\) &  99.85 &  91.33 & 95.50 & 95.18 & 90.13 &  100.00 & 59.46 & 83.33 & 54.11 & 83.59 & 83.73 & 77.30 & 98.35 & 96.50 & 98.00 & 78.05 & 80.15 & 78.50 & 98.20 & 86.38 \\
\rowcolor{lightblueAlpha}  DeeCLIP (ours) &  Blur \(\sigma =2\) & 98.00 & 76.27 & 85.80 & 82.32 & 82.21 & 99.45 & 57.72 & 85.00 & 49.09 & 76.58 & 85.91 & 70.75 & 88.90 & 83.75 & 88.20 & 72.05 & 71.40 & 72.00 & 89.60 & 79.74 \\
\rowcolor{lightblueAlpha}  DeeCLIP (ours) &  Blur \(\sigma =3\) & 90.20 & 75.44 & 76.08 & 68.73 & 83.75 & 93.05 & 56.30 & 83.33 & 47.72 & 60.61 & 73.26 & 54.20 & 73.35 & 66.25 & 72.25 & 59.15 & 58.25 & 59.10 & 75.60 & 69.82 \\ \midrule

 C2P-CLIP &  Average &  88.40 & \bf90.58 & \bf87.20 & \bf73.75& \bf93.44 & \bf91.54 & \bf83.75 & \bf94.21 & \bf53.77 & \bf74.18 & \bf74.49 & 57.02 & 12.92 & 39.83 & 14.37 & 40.85 & 42.50 & 38.57 & 18.08 & 61.55 \\ 
\rowcolor{lightblueAlpha}  DeeCLIP (ours) &  Average  &  \bf90.68 & 82.88 & 79.37 & 67.60 & 86.04 & 88.00 & 54.20 & 82.45 & 50.38 & 63.27 & 68.52 & \bf60.51 & \bf77.76 & \bf69.05 & \bf77.82 & \bf62.49 & \bf62.00 & \bf62.18 & \bf81.12 & \bf71.91  \\

\bottomrule
\end{tabular}
\end{adjustbox}  \vspace{-3mm}
\end{table*}

\begin{figure*}[t]
    \centering
    \includegraphics[width=\linewidth]{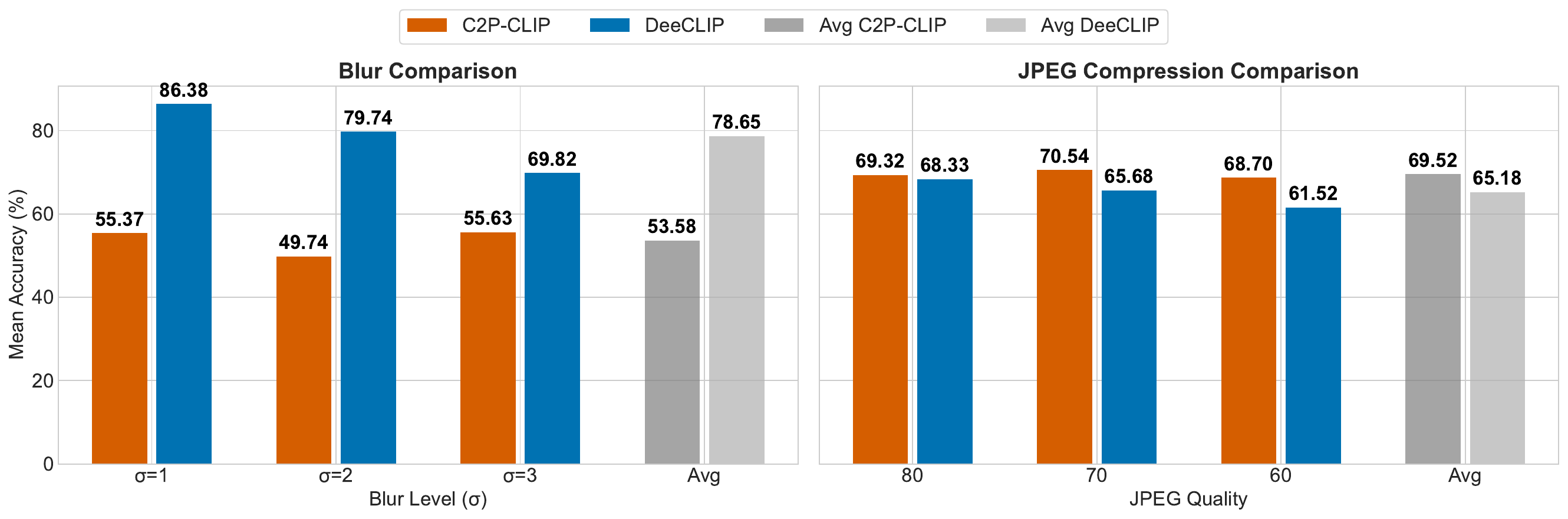}
    \caption{DeeCLIP's robustness under gaussian blur and JPEG compression, two common real-world degradations affecting AI-generated images. }
    \label{fig:deg} \vspace{-3mm}
\end{figure*}

\noindent\textbf{Robustness to Image Degradation.}
We evaluate the robustness of DeeCLIP and C2P-CLIP under two common image degradations: Gaussian blur and JPEG compression. These distortions simulate real-world challenges where image quality loss affects AI-generated content detection. Table~\ref{tab:perturb} presents the quantitative results, and performance trends are illustrated in Figure~\ref{fig:deg}.
Under Gaussian blur, performance decreases as severity increases, but DeeCLIP consistently outperforms C2P-CLIP. At mild blur ($\sigma=1$), DeeCLIP achieves 86.38\%, far exceeding C2P-CLIP’s 55.37\%. Even at severe blur ($\sigma=3$), DeeCLIP retains 69.82\% accuracy, while C2P-CLIP drops to 55.63\% (Figure~\ref{fig:deg}). Moreover, under JPEG compression, both methods degrade with lower quality, but DeeCLIP remains competitive. At q=80, DeeCLIP scores 68.33\%, close to C2P-CLIP’s 69.32\%. Even at q=60, DeeCLIP maintains 61.52\%, slightly lower than C2P-CLIP (68.70\%), showing its robustness to compression artifacts (Table~\ref{tab:perturb}). 

C2P-CLIP exhibits resilience to degradations in images generated by GAN-based models but struggles significantly when applied to images generated by diffusion models (DMs) under the same conditions. This discrepancy is particularly evident in categories such as LDM (200 steps) and DALL·E, where it achieves low average accuracies of 12.92\% and 18.08\% across all degradation types. In contrast, DeeCLIP performs more consistently under GAN-based and diffusion models (DMs), achieving a higher overall average accuracy of 71.91\%, compared to 61.55\% for C2P-CLIP, a gain of 10.36\% that underlines its robustness to real-world challenges. In summary, DeeCLIP consistently demonstrates higher resilience, particularly against spatial distortions, due to its feature fusion strategy, which preserves essential details. These results validate its effectiveness in real-world scenarios where image quality is often compromised.\\

\noindent\textbf{Ablation study.}
We conducted an ablation study to assess the impact of tuning the CLIP-ViT image encoder using the \acs{lora} technique, following the same training strategy with the 4-class ProGAN dataset. The experiment focused on evaluating average accuracy across both GAN and diffusion model datasets. Specifically, we compared the performance of using CLIP-ViT as a fixed backbone versus incorporating task-specific adaptation. Our results indicate that fine-tuning the CLIP-ViT image encoder with \acs{lora} significantly boosts detection accuracy, improving from 84.53\% with a fixed backbone to 89.00\%. This improvement highlights that while CLIP-ViT's pre-trained features provide general image representations, task-specific fine-tuning enhances its ability to identify subtle, model-specific artifacts in AI-generated images.

\vspace{-3mm}
\section{Conclusion}
\label{sec:conclusion}
\vspace{-3mm}
We proposed DeeCLIP, a novel and highly generalizable transformer-based framework for detecting AI-generated images. DeeCLIP builts on the strengths of the CLIP-ViT model by introducing three key innovations: (1) fine-tuning the CLIP-ViT image encoder using LoRA, a parameter-efficient adaptation technique, (2) integrating deep and shallow features through our DeeFuser module to enhance semantic alignment while preserving fine-grained details, and (3) incorporating triplet loss into the training process to refine the learned feature space, improving the separation between real and synthetic images. Our experimental evaluation on both GAN-based and diffusion-based datasets demonstrated that DeeCLIP achieved \ac{sota} generalization while maintaining significantly lower computational costs compared to conventional fine-tuning methods. Unlike prior approaches that relied on full fine-tuning or prompt-based tuning, DeeCLIP effectively balanced efficiency, adaptability, and robustness in detecting synthetic images. 
\vspace{-3mm}

\bibliographystyle{splncs04}
\bibliography{mybibliography}

\end{document}